# Exploring Parallelism in Learning Belief Networks

T. Chu and Y. Xiang
Dept. Computer Science, Univ. of Regina
Regina, Sask., Canada

## Abstract

It has been shown that a class of probabilistic domain models cannot be learned correctly by several existing algorithms which employ a single-link lookahead search. When a multi-link lookahead search is used, the computational complexity of the learning algorithm increases. We study how to use parallelism to tackle the increased complexity in learning such models and to speed up learning in large domains. An algorithm is proposed to decompose the learning task for parallel processing. A further task decomposition is used to balance load among processors and to increase the speed-up and efficiency. For learning from very large datasets, we present a regrouping of the available processors such that slow data access through file can be replaced by fast memory access. Our implementation in a parallel computer demonstrates the effectiveness of the algorithm.

## 1 INTRODUCTION

As the applicability of belief networks has been demonstrated in different domains, and many effective inference techniques have been developed, the acquisition of such networks from domain experts through elicitation becomes a bottleneck. As an alternative to manual knowledge acquisition, many researchers have actively investigated methods for learning such networks from data (Cooper & Herskovits 1992; Heckerman et al. 1995; Herskovits & Cooper 1990; Lam & Bacchus 1994; Spirtes et al. 1991; Xiang et al. 1997).

Since learning belief networks in general is NP-hard (Chickering et al. 1995), it is justified to use heuristic search in learning. Many algorithms developed use a scoring metric combined with a search procedure. In these algorithms, a single-link lookahead search is commonly adopted for efficiency. In a single-link lookahead search, consecutive network structures adopted differ by only one link. However, it has been shown that there exists a class of domain models termed pseudo-independent (**PI**) models which cannot be learned correctly by a single-link lookahead search (Xiang et al. 1996). One alternative for learning PI models is to use multi-link lookahead search (Xiang et al. 1997), where consecutive network structures may differ by more than one link. Increasing the number of links to lookahead, however, increases the complexity of learning computation.

In this work, we study parallel learning of belief networks. Parallel learning not only can be used to tackle the increased complexity during multi-link lookahead search, but also can speed up learning computation during single-link lookahead search in a large domain. Although parallel learning of rules have been studied (Cook & Holder 1990; Provost & Aronis 1996; Shaw & Sikora 1990), we do not realize other works on parallel learning of belief networks. Our study focuses on learning decomposable Markov networks (DMNs), although our result can be generalized to learning Bayesian networks. We study the parallelism using a message passing MIMD (multiple instruction multiple data) parallel computer.

We shall assume that readers are familiar with commonly used graph-theoretic terminologies such as cycle, connected graph, DAG, chordal graph, clique, junction tree (JT), sepset in a JT, I-map, etc. A *junction forest* (JF) $F$ of chordal graph $G$ is a set of JTs, each of which is a JT of one component of $G$.

The paper is organized as follows: In Section 2, we introduce PI models and multi-link lookahead search. In section 3, we propose parallel algorithms for learning belief networks. We also analyze the problems of load balancing and local memory limitation, and present our solutions. In section 4, we present our experimental results.

## 2 BACKGROUND

To make this paper self-contained, we give a brief introduction of PI models and multi-link lookahead search.

### 2.1 PSEUDO-INDEPENDENT MODELS

It has been shown (Xiang et al. 1997) that there exists a class of probability domain models where proper subsets of a set of collectively dependent variables display marginal independence. Examples of PI models are *parity* problems and *Modulus addition* problems (Xiang 1996). Several algorithms for learning belief networks have been shown being unable to learn correctly when the underlying domain model is a PI model. A simple PI model with four variables is shown in Table 1. More examples can be found in (Xiang et al. 1996).

Table 1: An example of PI models

| $(X_1, X_2, X_3, X_4)$ | $P(N)$ | $(X_1, X_2, X_3, X_4)$ | $P(N)$ |
|---|---|---|---|
| (0,0,0,0) | 0.0225 | (1,0,0,0) | 0.02 |
| (0,0,0,1) | 0.2025 | (1,0,0,1) | 0.18 |
| (0,0,1,0) | 0.005 | (1,0,1,0) | 0.01 |
| (0,0,1,1) | 0.02 | (1,0,1,1) | 0.04 |
| (0,1,0,0) | 0.0175 | (1,1,0,0) | 0.035 |
| (0,1,0,1) | 0.0075 | (1,1,0,1) | 0.015 |
| (0,1,1,0) | 0.135 | (1,1,1,0) | 0.12 |
| (0,1,1,1) | 0.09 | (1,1,1,1) | 0.08 |

It can be verified that $X_1$ and $X_4$ are conditionally independent given $X_2$ and $X_3$. In the subset $\{X_2, X_3, X_4\}$, each pair is marginally dependent, e.g., $P(X_2, X_3) \neq P(X_2)P(X_3)$, and is still dependent given the third, e.g., $P(X_2|X_3, X_4) \neq P(X_2|X_4)$. However, a special dependence relationship exists in the subset $\{X_1, X_2, X_3\}$. Although each pair is dependent given the third, e.g., $P(X_1|X_2, X_3) \neq P(X_1|X_2)$, $X_1$ and $X_2$ are marginally independent, i.e., $P(X_1, X_2) = P(X_1)P(X_2)$, so are $X_1$ and $X_3$. $\{X_1, X_2, X_3\}$ are said to be pairwise independent but collectively dependent. They form an *embedded* PI submodel. The minimal I-map of this model is shown in Figure 1 (a).

Suppose learning starts with an empty graph or structure (with all nodes but without any link). A single link lookahead search will not connect $X_1$ and $X_2$ since the two variables are marginally independent. Neither will $X_1$ and $X_3$ be connected. This results in the learned DMN structure in Figure 1 (b), which is incorrect. On the other hand, if we perform a double link search after the single-link search, which can effectively test whether $P(X_1|X_2, X_3) = P(X_1|X_2)$ holds, then the answer will be negative and the two links $(X_1, X_2)$

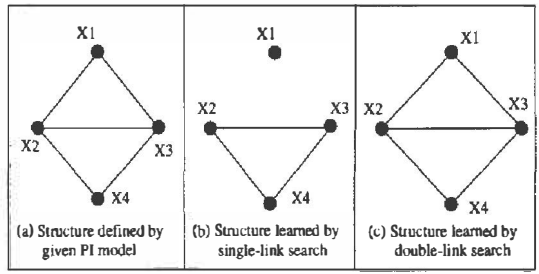

Figure 1: Comparison of learning results.

and $(X_1, X_3)$ will be added. The learned DMN structure is shown in Figure 1 (c).

### 2.2 A MULTI-LINK LOOKAHEAD SEARCH ALGORITHM

As our parallel learning algorithm is developed based on a multi-link lookahead search algorithm (Xiang et al. 1997), the latter is briefly introduced below.

**Algorithm** (Sequential)

Input: A dataset $D$ over a set $N$ of variables, a maximum size $\eta$ of clique, a maximum number $\kappa \leq \eta(\eta-1)/2$ of lookahead links, and a threshold $\delta h$.
begin
    initialize an empty graph $G = (N, E)$;
    $G' := G$;
    for $i = 1$ to $\kappa$, do
        repeat
            initialize the entropy decrement $dh' := 0$;
            for each set $L$ of $i$ links $(L \cap E = \phi)$, do
                if $G^* = (N, E \cup L)$ is chordal and $L$ is implied by a single clique of size $\leq \eta$, then
                    compute the entropy decrement $dh^*$;
                    if $dh^* > dh'$, then $dh' := dh^*$, $G' := G^*$;
            if $dh' > \delta h$, then $G := G'$, $done := false$;
            else $done := true$;
        until $done = true$;
    return $G$;
end

The search is structured into *levels* and the number of lookahead links is identical in the same level. Each level consists of multiple *passes*. Each pass at the same level tries to add the same number $i$ of links, that is, alternative structures that differ from the current structure by $i$ links are evaluated. For instance, level one search adds a single link in each pass, level two search adds two links, and so on. Search at each pass selects $i$ links that decrease the cross entropy maximally after testing all distinct and legal combinations of $i$ links. If the corresponding entropy decrement is significant enough, the $i$ links will be adopted and search continues at the same level. Otherwise, the next higher level of search starts.

Note that each intermediate graph is chordal as indicated by the *if* statement in the inner-most loop. The condition that $L$ is implied by a single clique $C$ means that all links in $L$ are contained in the subgraph induced by $C$. This requirement helps to reduce the search space.

## 3 PARALLEL LEARNING OF BELIEF NETWORKS

Learning a belief network using a single link lookahead search requires checking of $O(N^2)$ alternative structures before a link is added. In an m link lookahead search, $O(N^{2m})$ structures must be checked before $m$ links can be added. We view parallel learning as an alternative to tackle the increased complexity in multi-link search, as well as to speed up the single link search when the domain is large.

### 3.1 PARALLEL LEARNING ALGORITHMS

We extend Algorithm (Sequential) to parallel learning based on the following observation: at each pass of search, the exploration of alternative structures are coupled only through the current structure, i.e., given the current structure, tests of alternative structures are independent of each other. Hence the tests can be performed in parallel.

**Algorithm** (Manager-1)

Input: A dataset $D$ over a set $N$ of variables, a maximum
      size $\eta$ of clique, a maximum number $\kappa \leq \eta(\eta-1)/2$
      of lookahead links, the total number $n$ of explorers,
      and a threshold $\delta h$ for the cross entropy decrement.
begin
   send $D$, $N$ and $\eta$ to each explorer;
   initialize an empty graph $G = (N, E)$;
   $G' := G$;
   for $i = 1$ to $\kappa$, do
      repeat
         initialize the cross entropy decrement $dh' := 0$;
         partition all graphs that differ from $G$ by $i$ links
            into $n$ sets;
         send one set of graphs and $G$ to each explorer;
         for each explorer
            receive $dh^*$ and $G^*$;
            if $dh^* > dh'$ then $dh' := dh^*$, $G' := G^*$;
         if $dh' > \delta h$, then $G := G'$, $done := false$;
         else $done := true$;
      until $done = true$;
   send a termination signal to each explorer;
   return $G$;
end

As mentioned earlier, our study is performed in an environment where processors communicate through message passing only (vs. shared memory). We partition the processors as follows. One processor is designated as the search *manager* and the others are network structure *explorers*. The manager executes Algorithm (Manager-1). It is responsible for generating alternative graphs based on the current graph. It then partitions these graphs into $n$ sets and distributes one set to each explorer. Each explorer executes Algorithm (Explorer-1). It checks chordality for each graph received and computes the cross entropy decrement $dh^*$ for each valid chordal graph. It then chooses the best graph $G^*$ and reports $dh^*$ and $G^*$ to manager. Manager collects the reported graphs from all explorers, selects the best, and then starts the next pass of search.

**Algorithm** (Explorer-1)
begin
   receive $D$, $N$ and $\eta$ from the manager;
   repeat
      receive $G$ and a set of graphs from the manager;
      initialize $dh^* := 0$ and $G^* := G$;
      for each received graph $G' = (N, L \cup E)$, do
         if $G'$ is chordal and $L$ is implied by a single
            clique of size $\leq \eta$, then compute $dh'$ locally;
         if $dh' > dh^*$, then $dh^* := dh'$, $G^* := G'$;
      send $dh^*$ and $G^*$ to the manager;
   until termination signal is received;
end

Figure 2 illustrates the parallel learning process with two explorers and a dataset of four variables $u$, $v$, $x$ and $y$. Only a single-link search is performed for simplicity. Manager starts with an empty current graph in (a). It sends six alternative graphs in (b) through (g) to explorer 1 and 2. Explorer 1 checks graphs in (b), (c) and (d), selects the one in (b), and reports to manager. Explorer 2 reports the one in (e) to manager. After collecting the two graphs, manager chooses the one in (b) as the new current graph. It then sends graphs in (i) through (m). Repeating the above process, manager finally gets the graph in (n) and sends graphs in (o) and (p) to all explorers. Since none of them decreases the cross entropy significantly, manager chooses the graph in (n) as the final result and terminates explorers.

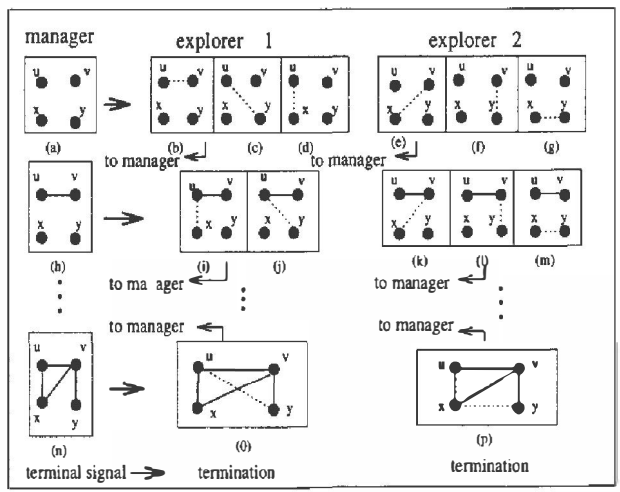

Figure 2: An example of parallel learning of DMN.

## 3.2   LOAD BALANCING

### 3.2.1   The Need of Load Balancing

Balancing load among processors is critical to the efficiency of parallel learning. In Algorithm (Manager-1), alternative graphs are evenly allocated to explorers. However, the amount of computation in checking each graph tends to switch between two extremes. If a graph is non-chordal, it is ignored immediately without having to compute the cross entropy decrement. For example, suppose the current graph is shown in Figure 3 (a). There are six graphs that differ from it by only one link. If any of the dotted links in (b) is added to (a), the resultant graph is non-chordal. Since the complexity of checking chordality is $O(|N|+|E|)$, where $N$ is the number of variables and E is the number of edges in the graph, the amount of computation is very small. On the other hand, if any of the dashed links in (c) is added to (a), the resultant graph is chordal. Since the complexity of computing cross entropy decrement by local computation is $O(n + \eta\,(\eta \log \eta + 2^\eta))$ (Xiang et al. 1997), where $n$ is the number of cases the dataset and $\eta$ is the maximum size of the cliques involved, the amount of computation is much larger. As a result, even job allocation may require significantly different amount of computation among explorers. As manager must collect reports from all explorers before a decision on the new current graph can be made, some explorers will be idle while other explorers are completing their jobs.

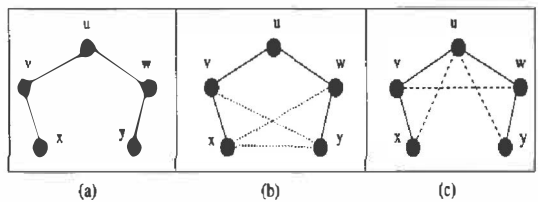

Figure 3: Two types of alternative structures.

Figure 4 shows the time taken by each of the six explorers in a particular search step. Explorer 1 takes much longer than others. This illustrates the needs for more sophisticated job allocation strategy in order to improve the efficiency of the parallel system.

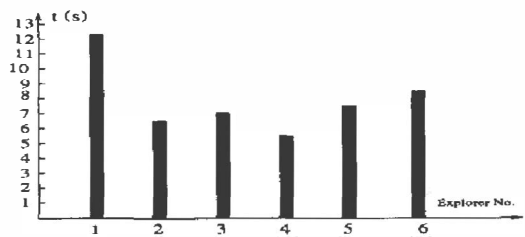

Figure 4: The time needed for each explorer.

### 3.2.2   Two-stage Loading Method

To improve load balancing, we modify Algorithms (Manager-1) and (Explorer-1) such that jobs are allocated in two stages. In the first stage, manager partitions alternative graphs evenly and distributes one set to each explorer. Each explorer checks the chordality for each graph received and reports to manager valid candidates (chordal graphs). Since the amount of computation for checking chordality is small, this stage can be completed quickly and the computation among explorers tends to be even. In the second stage, manager partitions all received graphs evenly and distributes one set to each explorer. Each explorer computes cross entropy decrement for each graph received. It then chooses the best graph $G^*$ and reports $dh^*$ and $G^*$ to manager. Manager collects the reported graphs, selects the best, and then starts the next pass of the search. Since all graphs are chordal in the second stage, the degree of load balance mainly depends on the variability of the sizes of the largest cliques.

## 3.3   MARGINAL SERVER

During learning, each explorer needs to extract marginal probabilities (marginals) for cliques from the dataset. If each processor must extract marginals by file access each time, the file system will become a bottleneck. One alternative is to compress the dataset and download one copy at each processor's local memory. This allows us to handle a dataset up to about 500MB in our parallel computer. However, when the size of dataset further increases, more sophisticated methods are needed.

According to the size of dataset, we partition the available processors into one manager, $n$ explorers and $m$ marginal *servers*. Each server's task is to compute marginals from the data stored in its local memory based on the request of explorers. Servers are connected logically into a pipeline indexed from 1 to m. The dataset is partitioned into $m+1$ sets. Each server stores one distinct set in its local memory. The last set is duplicated at each explorer's local memory. Algorithms (Manager-1) and (Explorer-1) are modified into Algorithms (Manager-2), (Explorer-2) and (Server).

The manger executes Algorithm (Manager-2). It performs data distribution as mentioned above. It then initializes an empty graph and starts the search. It generates alternative graphs based on the current graph, partitions into $m + n$ sets and distributes one set to each explorer and each server. It receives the valid candidates from explorers and servers, partitions them into $n$ sets, and send one to each explorer. It then collects the reported graphs from explorers, selects the best, sends a signal to each server, and starts the next pass of search.

**Algorithm** (Manager-2)

Input: A dataset $D$ over $N$ variables, a maximum size $\eta$ of clique, a maximum number $\kappa \leq \eta(\eta-1)/2$ of lookahead links, the total number $n$ of explorers, the total number $m$ of servers and a threshold $\delta h$.

begin
    partition $D$ into $m+1$ sets, send one distinct set to each server and broadcast the last set to explorers;
    initialize an empty graph $G = (N, E)$;
    $G' := G$;
    for $i = 1$ to $\kappa$, do
        repeat
            initialize the cross entropy decrement $dh' := 0$;
            partition all graphs that differ from $G$ by i links into $m + n$ sets;
            send one set of graphs and $G$ to each explorer and each server;
            for each explorer and server, do
                receive a set of valid graphs;
            partition all received graphs into n sets;
            send one set of graphs to each explorer;
            for each explorer
                receive $dh^*$ and $G^*$;
                if $dh^* > dh'$ then $dh' := dh^*$, $G' := G^*$;
            if $dh' > \delta h$, then $G := G'$, $done := false$;
            else $done := true$;
            send a signal to each server;
        until done = true;
    send a signal to each explorer and server;
    return $G$;
end

**Algorithm** (Explorer-2)

begin
    receive a set of data over a set $N$ of variables and a maximum size $\eta$ of clique;
    repeat
        receive $G$ and a set of graphs from the manager;
        initialize $dh^* := 0$ and $G^* := G$;
        for each received graph $G' = (N, L \cup E)$, do
            if $G'$ is chordal and $L$ is implied by a single clique of size $\leq \eta$, then mark it valid;
        send the valid candidates to manager;
        receive a set of graphs from manager;
        for each received graph $G' = (N, L \cup E)$, do
            for each clique involved in computing $dh'$,
                send the nodes of the clique to servers;
                compute marginal based on its local data;
                receive marginal from the server $m$;
                sum the two marginals up;
            compute the entropy decrement $dh'$ locally;
            if $dh' > dh^*$, then $dh^* := dh'$, $G^* := G'$;
        send $dh^*$ and $G^*$ to manager;
    until received signal;
end

Each explorer executes Algorithm (Explorer-2). It checks chordality for each graph received and reports to manager the chordal candidates. Next, it receives a set of chordal graphs from manager. For each graph received, it sends the nodes of each clique involved in computing the cross entropy decrement $dh$ to servers. After sending a request, it computes a sub-marginal based on its local data, receives a sub-marginal from the last server and sums them up. It then computes the cross entropy decrement $dh$. Finally, it reports the best graph $G^*$ to manager.

Each server executes Algorithm (Server). It does the same as an explorer for checking chordality. It then processes requests from explorers until a signal is received to start the next pass of search. After receiving a set of nodes (variables) from an explorer, it computes the sub-marginal using its local data, sums with the sub-marginal from its preceding server, and passes the sum to the next server or the requesting explorer.

**Algorithm** (Server)

begin
    receive a set of data over a set $N$ of variables and a maximum size $\eta$ of clique;
    repeat
        receive $G$ and a set of graphs from the manager;
        for each received graph $G' = (N, L \cup E)$, do
            if $G'$ is chordal and $L$ is implied by a single clique of size $\leq \eta$, then mark it valid;
        send the all valid candidates to manager;
        repeat
            receive a set of nodes from an explorer;
            compute the sub-marginal of the nodes received;
            if the server is not server 1, then
                receive the sub-marginal from its predecessor;
                sum the sub-marginal;
            if the server is not server m, then
                send the sub-marginal to the next server;
            else send the marginal to the requested explorer;
        until received signal;
    until received signal;
end

Since each server has to serve $n$ explorers, the processing of each server must be $n$ times as fast as an explorer. This means $nT_m = T_e$, where $T_m$ and $T_e$ are the computing time of each marginal server and each explorer, respectively. Let $|D_m|$ ($|D_e|$) be the size of local data at a server (explorer). $T_m$ and $T_e$ can be expressed as $T_m = k_d|D_m|$ and $T_e = k_g N + k_d|D_e|$, where $k_d$ and $k_g$ are coefficients, $N$ is the total number of variables in the domain, and $k_g N$ is the computation time to get the subgraph for computing the cross entropy decrement. Therefore, we have

$$nk_d|D_m| = k_g N + k_d|D_e|. \qquad (1)$$

Recall that we partition the dataset $D$ into $m+1$ sets,

$$|D| = m|D_m| + |D_e|. \qquad (2)$$

Solving equation 1 and 2 and $W' = m + n$, where $W = W' + 1$ is the total number of processors available, we get

$$n = \frac{W'(\alpha N + |D_e|)}{\alpha N + |D|},$$

$$m = \frac{W'(|D| - |D_e|)}{\alpha N + |D|},$$

$$|D_m| = \frac{\alpha N + |D|}{W'},$$

where $\alpha$ is a coefficient presented by $\alpha = k_g/k_d$, whose value is between 0.003 to 0.006 in our experimental environment. Furthermore, $|D_e|$ has to satisfy $|D_e| \leq M_d$, where $M_d$ is the maximum local memory available to store the data. For example, $T_e \approx 0.024 sec.$, $T_m \approx 0.005 sec.$, and $\alpha \approx 0.003$ in learning the $ALARM$ network with $n = 4$, $m = 4$, $|D_e| = 0.08MB$, and $|D_m| \approx 0.04MB$. If learning is performed on a large domain with a very large dataset, more marginal servers are needed. As an example, suppose $|D| = 100MB$, $N = 1000$, $W' = 30$ and $\alpha = 0.005$. If we choose $|D_e| = 20MB$, we obtain $n = 7$, $m = 23$, $|D_m| \approx 3.48MB$, where $n$ and $m$ are rounded to the nearest integers.

## 4 EXPERIMENTAL RESULTS

### 4.1 CONFIGURATION

The previous algorithms are implemented on an ALEX AVX Series 2 parallel computer with a MIMD distributed-memory architecture. It has 64 40$MHz$ processors each with 32 $MB$ local memory and can be directly linked to at most four others. Communication among processors are through message passing at 10 Mbps for simplex and 20 Mbps for duplex communication.

Message passing time increases as the number of links between the communicating processors and the length of the message. Table 2 shows the relation of the message passing time with the message length and the number of links between the communicating processors. It is important to link the processors such that the number of links involved in each message passing is minimized. Since each processor has up to four links and communication is to be performed among manager, explorers and servers, candidate topologies are a 2$D$ mesh and a ternary tree.

Table 2: Message passing time

| Length(bytes) | 256 | 1024 | 4096 | 16384 |
|---|---|---|---|---|
| 1 link | 0.015 | 0.016 | 0.023 | 0.096 |
| 2 links | 0.016 | 0.020 | 0.035 | 0.129 |
| 3 links | 0.017 | 0.022 | 0.044 | 0.125 |
| 7 links | 0.021 | 0.032 | 0.081 | 0.165 |
| 15 links | 0.030 | 0.057 | 0.160 | 0.241 |
| 31 links | 0.051 | 0.105 | 0.328 | 0.409 |

In a 2$D$ mesh with $W$ processors, the maximum number of links between any two processors is $D_{max} = 2(\sqrt{W} - 1)$. In a ternary tree with $W$ processors, the maximum number of links between manager (as the root of the tree) and an explorer is $T_{max} = \lceil \log_3^{(2W+1)} \rceil - 1$. The maximum number of links between an explorer and a server $2T_{max}$. In general, $T_{max}$ is smaller than $D_{max}$. For example, if $W = 25$, then $D_{max} = 8$ but $T_{max} = 3$. Therefore, the best topology is a ternary tree for our parallel learning algorithms. Figure 5 illustrates such a configuration. When there is no need for servers, all non-root processors are explorers.

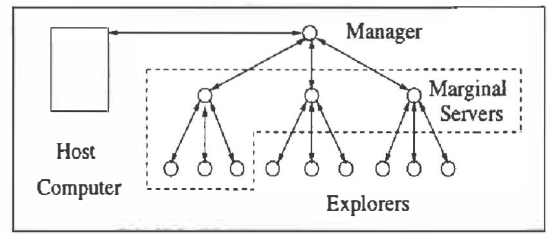

Figure 5: A ternary tree configuration

### 4.2 RESULTS AND PERFORMANCE ANALYSIS

Our experiments are intended to check the correctness of the parallel algorithms and the speed up through parallelism.

A network structure learned from a dataset generated from the *Alarm* network (Beinlich et al. 1989) is shown in Figure 6. The result obtained using the parallel algorithms is identical with that obtained using the sequential algorithm.

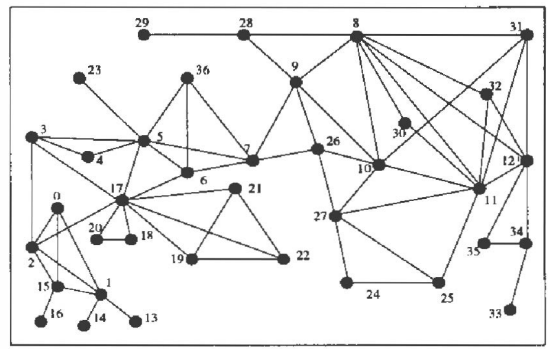

Figure 6: The learned Markov network: Alarm

We generated four control PI models and tested using the parallel algorithms with single-link and multi-link lookahead search. The model $PIM1$ has 26 variables and contains one PI sub-model of three variables. $PIM2$ and $PIM3$ have 30 and 35 variables, respectively. Each contains two PI sub-models each of which has three variables. $PIM4$ has 16 variables and contains a PI sub-model of four variables. The datasets of cases are generated by sampling these models with 20000, 25000, 30000 and 10000 cases, respectively.

For each dataset, our parallel algorithms were able to learn an approximate I-map of the control model. The network learned from $PIM3$ is shown in Figure 7. The two subsets of variables involved in the two PI sub-models are $\{x_6, x_8, x_9\}$ and $\{x_{14}, x_{15}, x_{16}\}$, respectively. Using a single link lookahead search, the dashed links (corresponding to the two PI sub-models) are missing (hence not an I-map) in the learning outcome. Using a triple-link lookahead search, the structure is learned correctly.

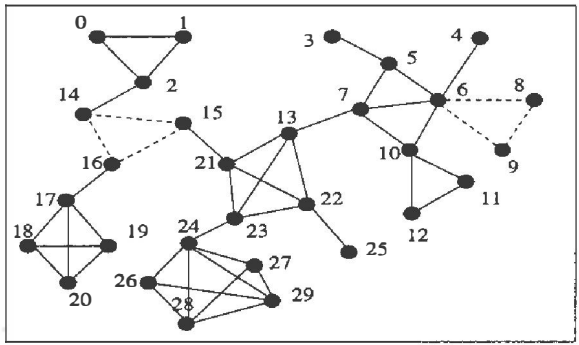

Figure 7: The structure learned from a PI model PIM3

Next, we present the performance result. The performance of a parallel program are commonly measured by *speed-up* (S) and *efficiency* (E). Given a particular task, let $T(1)$ be the execution time of a sequential program and $T(W)$ be that of a parallel program with $W$ processors. The two measurements are defined as $S = T(1)/T(W)$ and $E = S/W$.

In practice, the performance of a parallel program is affected by many factors. For our learning problem, the size of the input data is not trivial. Hence how the dataset is accessed by processors affects the performance significantly. Our experiments were intended to test the learning program using different data access methods. We also tested different job allocation methods which also affect the performance significantly.

For efficient $I/O$ and storage, the original dataset of cases are converted into a compressed frequency table where the value of each variable is represented by one byte. This file is used as the input to the learning program. In the parallel computer we used, file access by all processors must be performed through a special processor. The simplest way for a learning program to extract marginals from the data is to access the file directly for each marginal. This *file access* method was tested using a dataset of 10000 cases generated from the $ALARM$ network.

The sequential program (one processor) learned the network in 12800 seconds. The parallel program took 9260 seconds using 2 processors, 8512 seconds using 3 processors, and 8706 seconds using 4 processors. The speed-up is very small as the number of processors increases and the efficiency is very low. The speed-up with four processors was even lower than with three processors. This appears to be due to the bottleneck in file access and the corresponding increase in overhead. The result suggests that the file access method should be avoided due to the extensive data access needed during learning.

When the entire dataset can be loaded into the local memory (about 20 $MB$ is available in our environment) of each processor, loading the dataset to the memory is performed once for all and each marginal can be extracted directly from the memory. We refers to this as the *memory access* method. We conducted a comparison between even job allocation and two-stage job allocation using the memory access method.

Table 3: Results on even and two-stage job allocation

|   | Even loading | | | two-stage loading | | |
|---|---|---|---|---|---|---|
| n | time(s) | S | E | time(s) | S | E |
| 1 | 3160 | 1 | 1 | 3160 | 1 | 1 |
| 2 | 1750 | 1.81 | 0.903 | 1614 | 1.95 | 0.977 |
| 3 | 1201 | 2.63 | 0.877 | 1090 | 2.86 | 0.952 |
| 4 | 957 | 3.30 | 0.825 | 830 | 3.72 | 0.929 |
| 5 | 764 | 4.14 | 0.827 | 686 | 4.50 | 0.900 |
| 6 | 712 | 4.44 | 0.740 | 578 | 5.19 | 0.865 |
| 7 | 604 | 5.23 | 0.747 | 525 | 5.92 | 0.845 |
| 8 | 558 | 5.66 | 0.708 | 471 | 6.69 | 0.837 |
| 9 | 528 | 5.98 | 0.665 | 448 | 7.31 | 0.813 |
| 10 | 486 | 6.50 | 0.650 | 410 | 8.04 | 0.804 |
| 11 | 467 | 6.77 | 0.615 | 381 | 8.43 | 0.766 |
| 12 | 454 | 6.96 | 0.580 | 378 | 9.00 | 0.750 |

The experiment used a dataset of 10000 cases generated from the ALARM network. The result is shown in Table 3. Each row is the result obtained by using $n$ explorers as indicated in the first column. Columns 2 through 4 present the result obtained with even job allocation and columns 5 through 7 present the result obtained with two-stage job allocation.

Columns 3 and 6 show that as the number of explorers increases, the speed-up increases as well with either job allocation method. It demonstrates that our parallel algorithm can effectively reduce the learning time. This provides positive evidence that parallelism is an alternative to tackle the computational complexity in learning large belief networks.

Comparing column 3 with 6 and column 4 with 7, it can be seen that the two-stage allocation further speeds up the learning process and improves the efficiency compared with even job allocation. For example, when eight explorers are used, the speed-up is 5.66 and efficiency is 0.708 for even allocation, and 6.69 and 0.837 for two-stage allocation.

The result also shows a gradual decrease in efficiency as the number of explorers increases. This efficiency decrease is mainly due to the job allocation overhead. Manager must allocate job to each explorer sequentially at the beginning of each search step. Therefore, each explorer is idle after its report in the previous search step is submitted and before the next job is assigned to it.

Table 4: Results on learning PI models

| n | T, S, E | PIM1 | PIM2 | PIM3 | PIM4 |
|---|---|---|---|---|---|
| 1 | T(min.) | 262.4 | 868.6 | 3555.4 | 36584 |
| 12 | T(min.) | 26.8 | 89.3 | 352.2 | 3382 |
|  | S | 9.8 | 9.7 | 10.1 | 10.8 |
|  | E | 0.82 | 0.81 | 0.84 | 0.90 |
| 24 | T(min.) | 17.2 | 54.2 | 179.4 | 1735 |
|  | S | 15.3 | 16.0 | 19.8 | 21.1 |
|  | E | 0.64 | 0.67 | 0.83 | 0.88 |
| 36 | T(min.) | 12.5 | 37.7 | 124.5 | 1197 |
|  | S | 21.0 | 23.0 | 28.6 | 30.6 |
|  | E | 0.58 | 0.64 | 0.79 | 0.85 |

Table 4 lists the experimental result in learning the four $PI$ models mentioned above. Triple-link lookahead search is used for learning $PIM1$, $PIM2$ and $PIM3$, respectively. Six-link lookahead search is used for learning $PIM4$. The first column indicates the number of explorers used. Each row shows computation time, speed up or efficiency as indicated by the second column. Each of the last four columns shows the result for learning one PI model.

The second row shows the computation time in sequential learning. It increases from $PIM1$ to $PIM4$. This is because the increased size of the domain for $PIM1$ through $PIM3$ (26, 30, 35 variables) and the increased size of the dataset (20000, 25000, 30000). $PIM4$ used the most computation time because it has a $PI$ submodel of 4 variables, which requires six-link lookahead search.

For all models, the speed-up increases as more explorers are employed. On the other hand, when more explorers are used, $PIM1$ has the fastest decrease in efficiency and $PIM4$ has the slowest decrease with the other two models in-between. This is highly correlated with the increase of computation time from $PIM1$ to $PIM4$. This is because as the search space becomes larger, the number of alternative graphs to be explored in each job allocation becomes larger. The consequence is that the message passing overhead becomes less significant compared with the search time and hence the efficiency improves. This result shows that parallel learning are quite suited for tackling the increased computational complexity in learning PI models.

When the size of dataset is beyond the available local memory of each processor, we suggest the use of marginal servers. Comparison of using different number of servers was performed in learning $ALARM$ network. Figure 8 shows the speed-up comparison and Figure 9 shows the efficiency comparison. The vertical axis is labelled by S for speed-up or E for efficiency. The horizontal axis is labelled by $W = m + n$, where $m$ is the number of marginal servers and $n$ is the number of explorers. $D_e$ is the size of data stored in the local memory of each explorer, and $D_m$ is that of each server. The speed up is calculated using sequential learning with file access as this is considered the alternative when marginal server is not used.

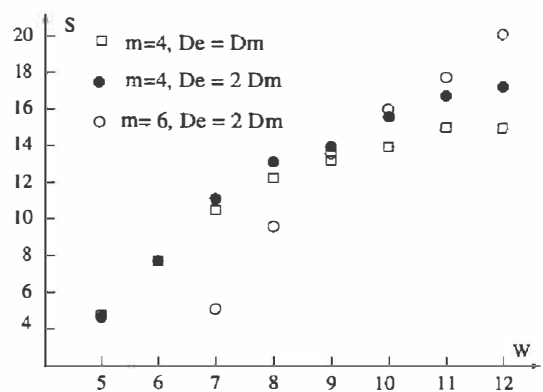

Figure 8: Speed-up by using marginal servers

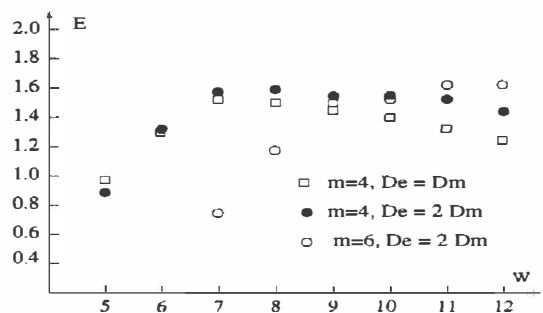

Figure 9: Efficiency by using marginal servers

In the Figure 9, the maximum efficiency is 1.528 for $m = 4$, $n = 3$ and $D_e = D_m$, 1.574 for $m = 4$, $n = 4$ and $D_e = 2D_m$, and 1.623 for $m = 6$, $n = 6$ and $D_e = 2D_m$. The corresponding speed-up is 10.69, 12.59 and 19.48, respectively. The speed up is more than the number of processors since marginal servers allow much faster memory access compared with the file access when a single processor is used.

## 5 REMARKS

We have studied parallelism in learning to tackle the increased computational complexity in learning belief networks in difficult domains (PI models) as well as in learning from large domains. Parallel algorithms were proposed that decompose the learning task such that multiple processors can be used without incurring additional error. In order to improve the efficiency of the parallel system, we proposed a two-stage job allocation method to handle the variation in computation time in searching different candidate networks. In order to overcome the bottleneck by file access, we proposed the parallel learning algorithm using marginal servers. This allows fast memory access of data when the size of the dataset is much larger than the local memory of each processor.

The parallel learning algorithms are implemented on an AVX Series 2 parallel computer with a MIMD distributed-memory architecture. Our experimental result showed that parallel learning can effectively speed up learning PI models as well as learning non-PI models in large domains.

### Acknowledgements

This work is supported by grants OGP0155425, CRD193296 from the Natural Sciences and Engineering Research Council of Canada, and by the Institute for Robotics and Intelligent Systems in the Networks of Centres of Excellence Program of Canada.